\documentclass[10pt,journal,compsoc]{IEEEtran}
\usepackage[nocompress]{cite}
\usepackage[pdftex]{graphicx}
\usepackage{balance}
\graphicspath{{figures/}}
\DeclareGraphicsExtensions{.pdf,.jpeg,.png}
\usepackage{amsmath,bm}
\usepackage{amsthm}
\usepackage{bbm}
\usepackage{algorithmic}
\usepackage{array}
\usepackage{url}
\usepackage[x11names]{xcolor}
\usepackage{tabulary,xfrac,enumitem}
\usepackage{booktabs}       
\usepackage{amsfonts}       
\usepackage{nicefrac}       
\usepackage{microtype}      
\usepackage{amssymb}
\usepackage{amsmath}
\usepackage{caption, subcaption}
\usepackage{multirow}
\usepackage{textcomp}
\usepackage{makecell}
\usepackage{colortbl}
\usepackage{pifont}
\usepackage{xspace}
\usepackage{subcaption}
\usepackage[ruled,linesnumbered]{algorithm2e}
\usepackage[pagebackref=true,breaklinks=true,colorlinks,citecolor=citecolor,bookmarks=false]{hyperref}
\definecolor{citecolor}{RGB}{34,139,34}

\begin{document}
\title{DPM++: Dynamic Masked Metric Learning for Occluded Person Re-identification}

\author{Lei Tan, Yingshi Luan, Pincong Zou, Pingyang Dai, Liujuan Cao
\IEEEcompsocitemizethanks{\IEEEcompsocthanksitem L. Tan, Y. Luan, P. Zou, P. Dai, and L. Cao are with Key Laboratory of Multimedia Trusted Perception and Efficient Computing, Ministry of Education of China, Xiamen University, 361005, P.R. China. Email: caoliujuan@xmu.edu.cn.}
\thanks{Corresponding author: Liujuan Cao}}

\markboth{Work on progress}%
{Shell \MakeLowercase{\textit{et al.}}: Bare Demo of IEEEtran.cls for Computer Society Journals}
\IEEEtitleabstractindextext{%
\begin{abstract}
Although person re-identification has achieved impressive progress in recent years, the common occlusion case caused by different obstacles remains an unsettled issue in real-world applications. The fundamental difficulty lies in the mismatch between incomplete occluded samples and holistic identity representations. Severe occlusion not only removes discriminative body cues but also introduces strong interference from background clutter and occluders, making conventional global metric learning unreliable. Existing methods mainly address this issue either by relying on extra pre-trained models to estimate visible parts for alignment or by constructing occluded samples through data augmentation to improve robustness. However, they still lack a unified framework that directly learns robust visibility-consistent matching under realistic occlusion patterns. In this paper, we propose DPM++, a Dynamic Masked Metric Learning framework for occluded person re-identification. The key idea of DPM++ is to learn an input-adaptive masked metric that dynamically selects reliable identity subspaces for each occluded instance, so that the matching process emphasizes visibility-consistent discriminative evidence while suppressing unreliable or occlusion-sensitive components. Built upon the original classifier-prototype space, DPM++ further introduces a CLIP-based two-stage supervision scheme, in which ID-level semantic priors are first learned from the text branch and then transferred into the classifier-prototype space used for final dynamic masked matching. To further strengthen the learning of the proposed masked metric, we introduce a saliency-guided patch transfer strategy to synthesize controllable and photo-realistic occluded samples during training. By exploiting real scene priors in the training data, this strategy exposes the model to realistic partial observations and provides richer supervision than conventional random erasing-based augmentation. In addition, occlusion-aware sample pairing and mask-guided optimization are developed to further improve the stability and effectiveness of the overall framework. Extensive experiments on multiple occluded and holistic person re-identification benchmarks demonstrate that DPM++ consistently outperforms previous state-of-the-art methods in both holistic and occlusion scenarios. 
\end{abstract}

}

\maketitle
\IEEEdisplaynontitleabstractindextext
\IEEEpeerreviewmaketitle

\IEEEraisesectionheading
{\section{Introduction}
\label{sec:introduction}}
Person re-identification (ReID) aims to match pedestrian instances across non-overlapping camera views and has been extensively studied due to its wide applications in intelligent surveillance and public security~\cite{zhu2023aaformer,tan2024partformer,li2023dc}. Benefiting from large-scale benchmarks and deep neural networks, recent ReID methods have achieved impressive progress under holistic settings, where most body regions are visible and discriminative appearance cues can be reliably captured~\cite{he2021transreid,zhang2023pha}. However, this assumption is frequently violated in real-world scenarios. In practice, pedestrians are often occluded by vehicles, other people, signboards, or background clutter, which severely corrupts identity evidence and causes a significant performance degradation for conventional ReID models. As a result, occluded person re-identification remains an important yet unresolved problem in practical deployment~\cite{tan2024occluded,xia2024attention,ye2024dynamic}.

\begin{figure}[t]
    \centering
    \includegraphics[width=1.0\columnwidth]{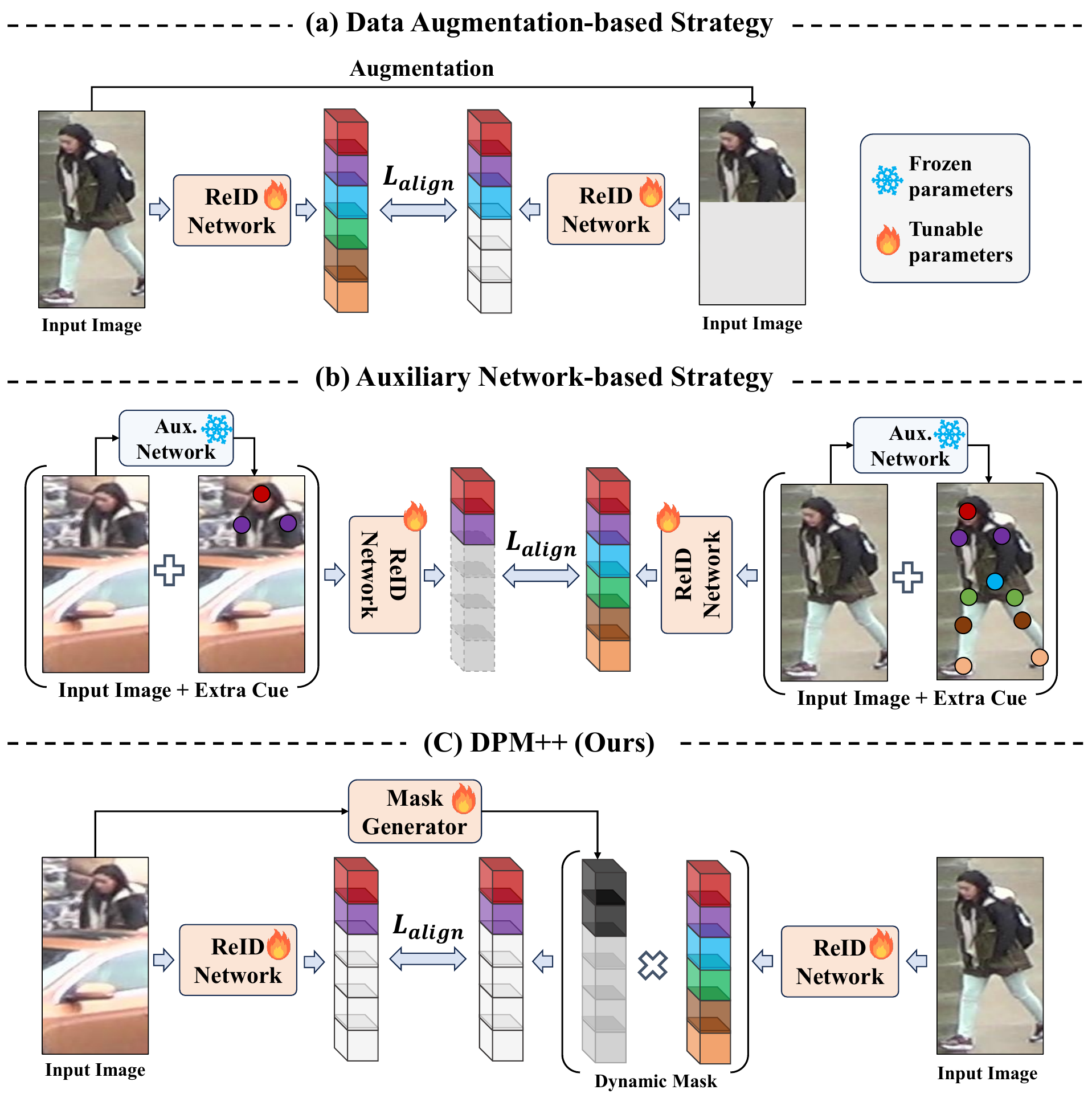}
    \caption{
    \textbf{Conceptual comparison of three paradigms for occluded person re-identification.}
    (a) Data augmentation-based methods improve robustness by enforcing feature consistency between the original sample and its augmented views.
    (b) Auxiliary network-based methods rely on extra cues, such as pose or visible-part information, to explicitly align occluded samples.
    (c) In contrast, DPM++ performs automatic partial-to-holistic alignment by dynamically generating an image-specific mask and selecting a visibility-consistent discriminative subspace, without requiring extra body-part supervision or pre-trained auxiliary networks.
    }
    \label{fig:motivation}
\end{figure}

The core difficulty of occluded person re-identification lies in the mismatch between occluded samples and holistic identity representations. Under severe occlusion, only a subset of identity-related patterns remains visible, while conventional ReID models are usually trained to compare global features in a fixed embedding space. Such global matching is vulnerable to missing body parts, spurious background responses, and occluder interference, making the learned metric unreliable when visual evidence becomes incomplete. Existing works mainly attempt to alleviate this issue from two directions, as shown in Figure~\ref{fig:motivation} (a) and (b). One line of work introduces extra pre-trained models~\cite{wang2020high,gao2020pose,wang2022pose}, such as human parsing, pose estimation, or body clue extraction, to estimate visible parts and assist feature alignment. Although effective in some cases, these methods often suffer from domain gaps between auxiliary models and ReID data, and their performance can be sensitive to inaccurate visible-part prediction. Another line of work enhances robustness through occlusion-oriented data augmentation by constructing occluded samples during training~\cite{zhong2020random,wang2022feature,chen2021occlude}. While such strategies improve data diversity, many of them still rely on random erasing-style corruption, whose content and location differ substantially from realistic occlusion patterns observed in practice.

These limitations indicate that occluded ReID should not be treated merely as a harder version of holistic ReID, nor as a problem that can be solved solely by auxiliary visible-part estimation or random corruption simulation. Instead, it should be revisited as a partial-to-holistic matching problem: given an incomplete visual observation, the model must determine which identity evidence remains reliable and compare it against a more complete identity anchor in a visibility-consistent manner. From this perspective, the key challenge is not only to suppress occlusion noise, but also to learn a metric that dynamically adapts to the visibility condition of each input instance.

A natural way to address this challenge is to introduce prototypes as holistic anchors and perform matching in a dynamically selected subspace~\cite{tan2022dynamic}. Prototype-based matching has shown strong potential in occluded ReID, since it provides a relatively stable reference against which incomplete observations can be compared. In particular, DPM~\cite{tan2022dynamic} demonstrates that dynamic prototype masking offers an effective way to reformulate occluded matching as subspace selection in the prototype space. Nevertheless, purely visual prototypes are still learned from training images and therefore inevitably inherit the bias of occlusion, background contamination, and data imbalance. Consequently, under severe occlusion, visual prototypes may not always serve as sufficiently complete or robust identity anchors. This limitation motivates us to move beyond purely visual supervision and seek stronger semantic priors for prototype learning.

In this paper, we propose \textbf{DPM++}, a \textbf{Dynamic Masked Metric Learning} framework for occluded person re-identification as shown in Figure~\ref{fig:motivation} (c). The central idea of DPM++ is to learn an input-adaptive masked metric that dynamically selects reliable identity subspaces for each occluded instance, so that the matching process emphasizes visibility-consistent discriminative evidence while suppressing unreliable or occlusion-sensitive components. Different from directly replacing the original DPM prototypes with text embeddings, DPM++ retains the task-oriented classifier-prototype space for final dynamic masked matching, while introducing a CLIP-based two-stage supervision scheme to enhance prototype learning. Specifically, ID-level semantic priors are first learned from the text branch, and then transferred into the classifier-prototype space used in the second stage. In this way, the final classifier prototypes become more holistic, semantically informed, and robust under severe occlusion, while remaining fully compatible with the original DPM-style masking mechanism.

This design offers two important advantages. First, the text branch provides higher-level semantic priors that are less sensitive to local corruption caused by occlusion. Compared with prototypes learned purely from occluded visual observations, the semantics learned from CLIP offer a more stable supervisory signal for identity representation learning. Second, by transferring such semantic priors into the classifier-prototype space, DPM++ preserves the strengths of the original DPM formulation, i.e., performing final masked matching in a task-oriented visual prototype space rather than directly relying on text features as the matching target. Therefore, DPM++ does not simply learn a stronger embedding, but learns how to perform dynamic partial-to-holistic matching in a semantically enhanced prototype space.

To fully exploit the potential of the proposed masked metric, we further introduce a \emph{saliency-guided patch transfer} mechanism as a realistic occlusion synthesis strategy during training. Different from conventional random erasing-based augmentation, this strategy explicitly exploits real scene priors already present in the training data. By separating salient identity patches from occlusion-related patches and recombining them under occlusion-aware pairing, we generate controllable and photo-realistic occluded samples that better reflect practical occlusion patterns. More importantly, these synthesized samples provide richer partial-observation supervision for DPM++, exposing the model to realistic occlusion variations and substantially strengthening the learning of the proposed dynamic masked metric. Therefore, in our framework, realistic occlusion synthesis is not an isolated augmentation trick, but a key component that improves the quality of visibility-aware metric learning.

Overall, DPM++ establishes a unified framework that connects realistic occlusion synthesis during training, CLIP-based semantic prior learning, and dynamic masked matching during inference. The first component enriches the model with realistic partial observations, the second transfers semantic guidance into the classifier-prototype space, and the third enables robust partial-to-holistic identity comparison at test time. In this sense, DPM++ closes two important gaps in occluded ReID: the gap between synthetic and realistic occlusion during training, and the gap between partial observations and holistic identity representations during matching. Extensive experiments on both occluded and holistic ReID benchmarks demonstrate that the proposed framework consistently improves robustness under realistic occlusion scenarios while preserving strong performance in standard holistic settings.

This paper is a substantial extension of our previous conference works, including Dynamic Prototype Mask (DPM)~\cite{tan2022dynamic} and Saliency-Guided Patch Transfer (SPT)~\cite{tan2024occluded}. The present work goes significantly beyond simply combining these two methods. First, while DPM formulates occluded person re-identification as dynamic prototype masking in a purely visual space, this paper generalizes it into a unified dynamic masked metric learning framework, namely DPM++, which explicitly addresses partial-to-holistic matching under occlusion. Second, instead of directly replacing visual class prototypes with text prototypes, we introduce a CLIP-based two-stage supervision scheme that learns ID-level semantic priors from the text branch and transfers them into the classifier-prototype space used for final masked matching. Third, while SPT was originally proposed as an independent data-driven augmentation strategy, in this work, it is seamlessly integrated into DPM++ as a realistic occlusion synthesis mechanism that strengthens the learning of dynamic masked metrics by providing controllable and photo-realistic partial observations during training. Finally, compared with the conference versions, this paper presents a more comprehensive formulation, an improved optimization framework, and substantially more extensive experiments and analyses, resulting in a stronger performance and more detailed ablation.

The main contributions of the paper are summarized as follows:
\begin{itemize}
    \item We propose DPM++, a novel Dynamic Masked Metric Learning framework for occluded person re-identification, which enables automatic partial-to-holistic alignment without relying on extra pre-trained models. By incorporating CLIP-based semantic supervision into the classifier-prototype space as a semantic anchor, DPM++ further enhances the robustness of the learned identity anchors for dynamic masked matching under occlusion.

    \item We develop a saliency-guided patch transfer strategy to synthesize controllable and photo-realistic occluded samples during training, which strengthens the learning of the proposed masked metric learning by exposing the model to more realistic partial observations.

    \item Extensive experiments on multiple occluded and holistic person re-identification benchmarks demonstrate that DPM++ consistently outperforms strong baselines and previous state-of-the-art methods, showing superior robustness and generalization under realistic occlusion scenarios.
\end{itemize}

The rest of the paper is organized as follows. Section II briefly reviews some related works on holistic and occluded person re-identification. The proposed DPM++ is detailed in Section III. The experiments are discussed and analyzed in Section IV. Finally, conclusions are drawn in Section V.

\begin{figure*}[t]
\centering
\includegraphics[width=2.0\columnwidth]{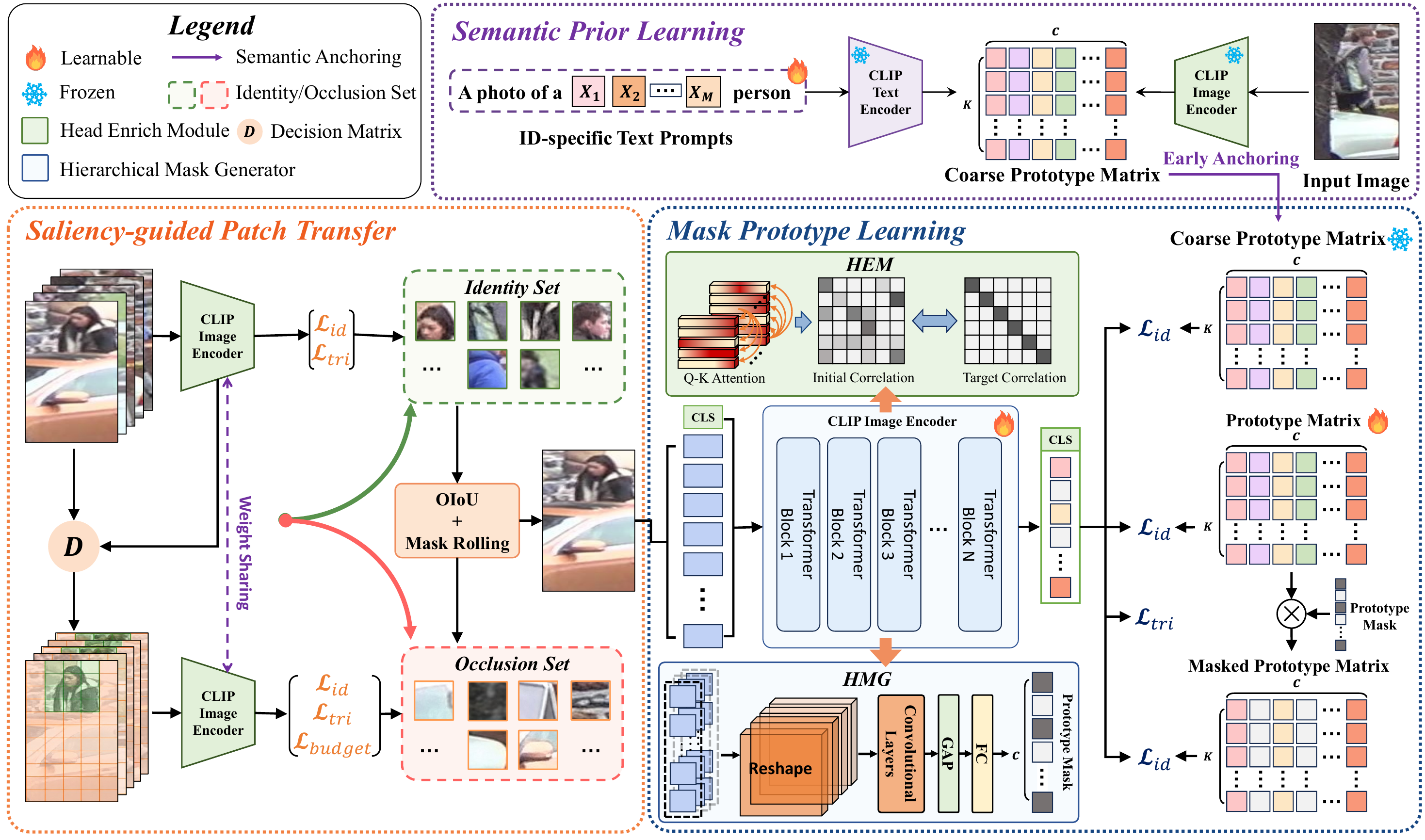}
\caption{
\textbf{Overall framework of DPM++.}
DPM++ consists of saliency-guided patch transfer and mask prototype learning, with an early semantic anchoring strategy introduced to stabilize prototype optimization. 
SPT decomposes training images into identity and occlusion sets and recombines them to synthesize realistic occluded samples. 
Mask prototype learning then predicts an input-specific prototype mask and performs visibility-consistent matching in the prototype space. 
The coarse prototype matrix learned from CLIP-based ID-specific prompts provides semantic anchoring for the learnable classifier-prototype space. 
HEM and HMG further improve the diversity of CLS representations and the reliability of prototype mask estimation.
}
\label{fig:overview}
\end{figure*}

\section{Related Works}
\subsection{Holistic Person Re-Identification}
Person re-identification (ReID) aims to match pedestrian instances across non-overlapping camera views. Early studies mainly relied on handcrafted descriptors~\cite{ma2014covariance,yang2014salient,liao2015person} together with carefully designed metric learning strategies~\cite{zheng2012reidentification,koestinger2012large}. With the rapid development of deep learning, feature representation learning has become the dominant paradigm in person ReID as well as many other vision tasks~\cite{he2017mask,mou2020plugnet,chen2021e2net,peng2021knowledge,zhang2021towards}. 

Among CNN-based methods, Luo \emph{et al}.~\cite{luo2019bag} propose the BN-Neck structure, which establishes a strong and widely adopted baseline for holistic person ReID. Chen \emph{et al}.~\cite{Chen_2019_Mixed} further introduce high-order attention to capture richer contextual dependencies, while Zheng \emph{et al}.~\cite{zheng2019joint} integrate discriminative and generative learning into a unified framework. Besides global feature learning~\cite{luo2019bag,zheng2019joint,ye2021deep}, part-based representation learning has also been extensively explored to provide fine-grained pedestrian descriptions. Representative methods such as PCB~\cite{sun2018beyond}, MGN~\cite{wang2018learning}, and Pyramid~\cite{zheng2019pyramidal} divide the input image or feature map into multiple parts to enhance local discriminability.

More recently, transformer-based architectures have shown strong potential in person ReID. TransReID~\cite{he2021transreid} first introduces the Vision Transformer (ViT) into ReID and demonstrates its superiority in modeling long-range dependencies. Subsequent iterations, such as Partformer~\cite{tan2024partformer}, AAFormer~\cite{zhu2023aaformer}, and DC-Former~\cite{li2023dc}, further explored fine-grained representations on the ViT structure. Recent works further extend holistic ReID toward more general and semantically enriched settings. For example, CLIP-ReID~\cite{li2023clip} further introduced the CLIP-based ViT encoder with text alignment for ReID task. PromptSG~\cite{yang2024pedestrian} leverages language-guided prompts to improve person ReID by injecting semantic guidance into representation learning, and Instruct-ReID~\cite{he2024instruct} reformulates multiple ReID settings into an instruction-driven framework, showing that language can provide a unified interface for retrieval under diverse conditions. CLIMB-ReID~\cite{yu2025climb} further explores a hybrid CLIP-Mamba architecture for person ReID and demonstrates the effectiveness of CLIP-based representations in this task. In addition, Pose2ID~\cite{yuan2025poses} proposes a training-free feature centralization framework that aggregates same-identity features to improve representation stability and reports strong adaptability across standard, cross-modality, and occluded ReID settings. ChatReID~\cite{niu2025chatreid} pushes this trend further by investigating open-ended interactive person retrieval in a text-side-dominated framework, highlighting the growing potential of language-guided person ReID paradigms. Despite these advances, most holistic ReID methods still assume that pedestrian bodies are largely visible, and therefore often suffer significant performance degradation in real-world scenarios with frequent occlusions.

\subsection{Occluded Person Re-Identification}
Occluded person re-identification highlights the weakness of holistic ReID methods under incomplete observations. The key challenge lies in the fact that severe occlusion destroys discriminative body information and introduces substantial interference from obstacles, making global feature comparison unreliable. Existing occluded ReID methods can be roughly categorized into \emph{model-driven} and \emph{data-driven} approaches.

Model-driven methods mainly focus on learning visibility-aware representations or alignment strategies. Early works attempt to suppress occlusion noise and directly extract global features from visible regions. Zhuo \emph{et al}.~\cite{zhuo2018occluded} introduce an auxiliary occluded/non-occluded classification task, while Chen \emph{et al}.~\cite{chen2021occlude} combine occlusion augmentation with attention mechanisms to better capture informative body regions. To further alleviate the mismatch caused by missing parts, some methods exploit structural priors or feature completion strategies. Wang \emph{et al}.~\cite{wang2020high} employ high-order relations and a human-topology graph based on pose estimation to propagate information from visible regions to invisible ones. Hou \emph{et al}.~\cite{hou2021feature} propose a feature completion module to infer occluded features from long-range spatial context. Another line of work follows the fine-grained matching paradigm and explicitly aligns visible parts. Miao \emph{et al}.~\cite{miao2019pose} propose PGFA to disentangle visible part information from occlusion noise using pose landmarks, Gao \emph{et al}.~\cite{gao2020pose} introduce visible part matching with pose-guided attention, and Li \emph{et al}.~\cite{li2021diverse} further exploit prototype learning with transformers to discover fine-grained body parts without extra networks. DPM~\cite{tan2022dynamic} moves beyond explicit body-cue supervision by formulating occluded ReID as dynamic prototype masking in a purely visual space, enabling automatic alignment without relying on extra pre-trained networks. More recent methods continue to improve robustness under partial observations. FPC~\cite{ye2024dynamic} has been introduced to reduce contamination from occluders, and KPR~\cite{somers2024keypoint} further addresses multi-person ambiguity in crowded and heavily occluded boxes with semantic keypoint prompts.

Data-driven methods improve robustness by augmenting training data with synthesized occlusion patterns. Huang \emph{et al}.~\cite{huang2018adversarially} construct adversarially occluded samples for data augmentation. Chen \emph{et al}.~\cite{chen2021occlude} design a prior-guided occlusion augmentation strategy, and Wang \emph{et al}.~\cite{wang2022feature} further develop a feature erasing and diffusion framework. More recently, Xia \emph{et al}.~\cite{xia2024attention} propose an Attention Disturbance and Dual-Path Constraint Network, which introduces disturbance masks to simulate more realistic occlusion interference and improves attention generalization through dual-path constraints. However, existing data-driven approaches are still closely related to the random erasing paradigm~\cite{zhong2020random}, and thus remain insufficiently sensitive to the content and spatial location of real-world occluders. Consequently, the generated samples often exhibit a noticeable gap from practical occlusion cases. Saliency-Guided Patch Transfer (SPT) moves one step further by synthesizing controllable and photo-realistic occluded samples through saliency-guided identity/occlusion decomposition and occlusion-aware pairing~\cite{tan2024occluded}. Compared with previous methods, our work not only inherits the advantage of realistic occlusion synthesis during training, but also reformulates occluded ReID as dynamic prototype-space matching. Moreover, instead of directly using text prototypes as final matching targets, DPM++ introduces CLIP-based coarse semantic priors to anchor the learnable classifier-prototype space.

\section{The Proposed Method}

\subsection{Overview}
\label{sec:overview}

The overall framework of \textbf{DPM++} is shown in Fig.~\ref{fig:overview}. 
DPM++ addresses occluded person re-identification from the perspective of semantically anchored mask prototype learning. 
Unlike conventional person ReID methods that directly compare complete image features in a shared embedding space, the proposed framework is explicitly designed for the realistic case where the input only provides partial, visibility-biased, and often contaminated information. 
The central idea is to jointly answer two tightly related questions of occluded matching: 
\emph{(i)} how to construct realistic occluded samples for training, and 
\emph{(ii)} how to perform visibility-aware matching between incomplete occlusion input query and diversity gallery. 
To this end, DPM++ integrates two tightly coupled core components, namely \emph{saliency-guided patch transfer} and \emph{mask prototype learning}, while an early semantic anchoring strategy is further introduced to stabilize prototype optimization from a semantic perspective.

More specifically, the first component, saliency-guided patch transfer (SPT), reconstructs the training regime for occluded ReID. 
Instead of applying random erasing or hand-crafted corruptions, DPM++ explicitly decomposes each training sample into an identity set and an occlusion set, and then recombines them under OIoU-guided mask rolling to synthesize photo-realistic occluded samples. 
As a result, the model is repeatedly exposed to realistic occluded samples whose occlusion content and spatial layout are inherited from the original training data. 
The second component, mask prototype learning, performs the final matching in a prototype space. 
Given an input image, the framework extracts a CLS feature, predicts an input-specific prototype mask, and dynamically matches the input against both a learnable prototype matrix and its masked counterpart. 
Meanwhile, a coarse prototype matrix learned from CLIP-based ID-specific text prompts is used as an early anchoring mechanism, providing semantically informed identity priors for the learnable prototype space. 
In this way, DPM++ unifies realistic occlusion synthesis, semantic anchoring, and dynamic prototype matching within a single framework.

Formally, given an input image $I \in \mathbb{R}^{H \times W \times 3}$, we first split it into $D$ non-overlapping image patches and flatten each patch into a token. 
Assume the patch size is $P$ and the stride is $s_d$, then the number of patches is
\begin{equation}
D =
\left\lfloor \frac{H-S_p}{s_d}+1 \right\rfloor
\left\lfloor \frac{W-S_p}{s_d}+1 \right\rfloor .
\label{eq:num_patch}
\end{equation}
After linear projection $\mathcal{F}(\cdot)$, a learnable class token $x_{cls}$ is prepended to the patch sequence. 
Following ViT-based person ReID, position embedding $\mathcal{P}$ and camera embedding $\mathcal{C}$ are further added:
\begin{equation}
z_0
=
[x_{cls}^{0}; \mathcal{F}(x_1^0); \mathcal{F}(x_2^0); \cdots; \mathcal{F}(x_D^0)]
+ \mathcal{P} + \lambda \mathcal{C},
\label{eq:token_init}
\end{equation}
where $\lambda$ is a hyper-parameter balancing the camera embedding. 
The initialized sequence $z_0$ is then fed into a stack of transformer blocks. 
The resulting CLS token is used as the global identity representation, while hierarchical patch features are exploited to predict an image-specific prototype mask. 
As shown in Fig.~\ref{fig:overview}, the final CLS representation is jointly supervised by three prototype spaces: a frozen \emph{coarse prototype matrix}, a learnable \emph{prototype matrix}, and an instance-specific \emph{masked prototype matrix}. 
These three branches respectively contribute semantic anchoring, standard identity discrimination, and visibility-consistent masked matching.

\begin{figure}[t]
    \centering
    \includegraphics[width=0.90\columnwidth]{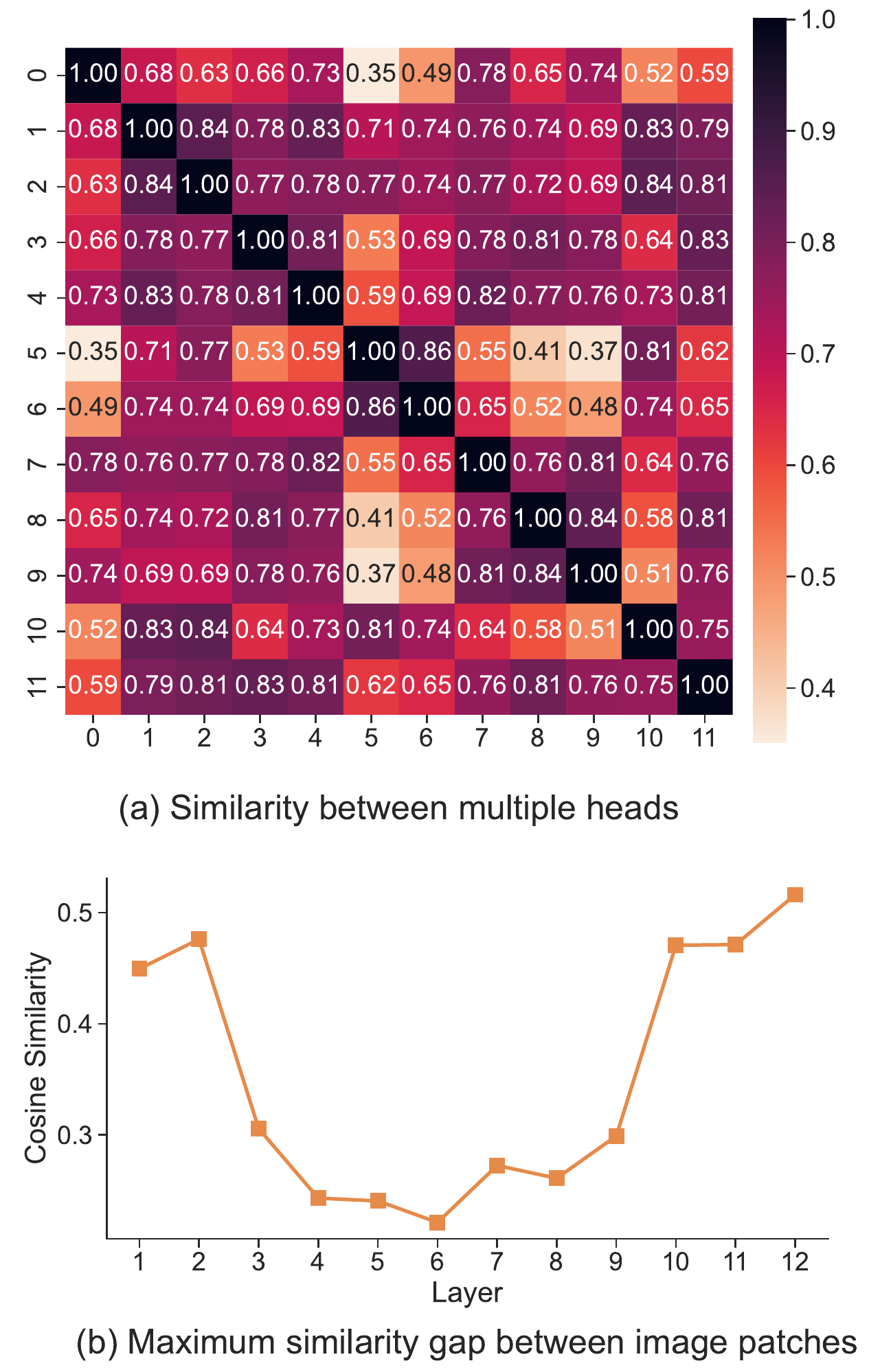}
    \caption{
    \textbf{Motivation for HEM and HMG.}
    (a) Cross-correlation matrix between attention maps of different heads in the last transformer block. 
    The high off-diagonal correlations indicate that different heads tend to attend to similar visual patterns, motivating the Head Enrich Module. 
    (b) Maximum similarity between image patches across transformer layers. 
    The reduced token discriminability in deeper layers motivates the Hierarchical Mask Generator, which aggregates multi-level features for reliable prototype mask estimation.
    }
    \label{fig:ori_heat}
    \vspace{-1em}
\end{figure}

\subsection{Saliency-Guided Patch Transfer}
\label{sec:spt}

The first core component of DPM++ is \emph{saliency-guided patch transfer} (SPT), whose role is to construct realistic occluded samples for training. 
Existing data-driven augmentation methods~\cite{chen2021occlude,wang2022feature,xia2024attention} for occluded ReID are largely rooted in the random erasing paradigm~\cite{zhong2020random}, where random noise or manually selected patches are used to cover image regions. 
Although such strategies can increase data diversity, the generated occlusions often exhibit substantial discrepancies in both content and location compared with real-world occlusion patterns. 
As pointed out in SPT, practical occlusions are highly structured: obstacles come from meaningful scene elements such as cars, signboards, or other persons, and their locations are constrained by the target person’s geometry and scale. 
Therefore, directly synthesizing occlusions from real scene patches is more natural than relying on arbitrary corruption. 

DPM++ adopts this philosophy and integrates SPT into the unified training pipeline. 
Given a training image, SPT first decomposes it into two subsets: an \emph{identity set}, which contains salient target-person patches, and an \emph{occlusion set}, which contains background or obstacle patches. 
Then, these two subsets are recombined under an OIoU-guided mask rolling strategy to generate realistic occluded samples. 
These synthesized samples are not merely auxiliary augmentations, they are the actual realistic occluded samples used to train the subsequent mask prototype learning branch. 
Therefore, SPT serves as the data construction branch of DPM++, bridging the gap between synthetic corruption and real occlusion. 

\subsubsection{Salient Patch Selection}

The first step of SPT is to separate pedestrian regions from background/occluding regions. 
Inspired by dynamic ViT-style token selection~\cite{rao2021dynamicvit,meng2022adavit}, SPT learns a decision matrix that predicts the importance of each patch in an end-to-end manner. 
Let $z_l$ denote the patch-token representation of the $l$-th transformer block after removing the class token. 
Following the SPT design, multi-level features are concatenated and passed through a decision layer to obtain a saliency mask:
\begin{equation}
M
=
\sigma\Big(
W^{p}\operatorname{Concat}(z_0,z_1,z_2,\dots,z_L)
\Big),
\label{eq:sps_decision}
\end{equation}
where $\sigma(\cdot)$ is the sigmoid function, $W^p$ is the learned decision matrix, and $L$ is the number of transformer blocks. 

The resulting saliency mask is then used in a second forward inference to filter out background patches:
\begin{equation}
Z_0
=
[z_0^{cls}; M \odot z_0].
\label{eq:sps_second}
\end{equation}
Patches associated with high saliency responses are assigned to the identity set, while the suppressed patches form the occlusion set. 
Notably, this decomposition is learned solely from identity discrimination signals, without any extra supervision from pose estimation, human parsing, or foreground segmentation. 
In other words, the network spontaneously learns which patches are identity-relevant and which patches are occlusion/background-related. 
This is particularly important in our setting, because the decomposition should be driven by ReID objectives rather than by a generic segmentation prior. 

\subsubsection{OIoU-Guided Patch Recombination}

Once the identity and occlusion sets are constructed, the next step is to recombine them into realistic occluded samples. 
A naive random combination may produce implausible overlap or severe mismatch in scale. 
To address this issue, DPM++ employs the \emph{Occlusion-Aware Intersection over Union} (OIoU) criterion proposed in SPT. 
Given a target instance mask $M_i$ from sample $I_i$ and a candidate mask $M_j$ from sample $I_j$, the standard IoU is given by:
\begin{equation}
IoU(M_i,M_j)=\frac{M_i\cap M_j}{M_i\cup M_j}.
\label{eq:iou}
\end{equation}
However, standard IoU treats the two masks symmetrically and tends to favor large-area candidates. 
To emphasize effective occlusion with compact support, OIoU replaces the denominator with the candidate mask:
\begin{equation}
OIoU(M_i,M_j)=\frac{M_i\cap M_j}{M_j}.
\label{eq:oiou_full}
\end{equation}
This criterion naturally prefers candidates that can cover the target region efficiently without introducing unnecessarily large occlusion areas. 

To further improve spatial compatibility, we apply a horizontal mask rolling strategy. 
The target mask $M_i$ is rolled with a small stride, and the maximum OIoU score over rolled positions is used for candidate selection. 
This strategy encourages the selected candidate to exhibit a scale more compatible with the target person, thereby avoiding unrealistic recombinations. 
The final synthesized token sequence is given by:
\begin{equation}
\begin{split}
Z_0^i &= [z_0^{cls}; M_j \odot z_0^i + (1-M_j)\odot z_0^j], \\
\text{s.t.}\quad
&OIoU(M_i,M_j)\ge \alpha_1, \\
&\max \big(OIoU(\operatorname{Roll}(M_i),M_j)\big)\ge \alpha_2,
\end{split}
\label{eq:spt_recombine}
\end{equation}
where $\alpha_1$ and $\alpha_2$ are thresholds controlling candidate selection. 
The synthesized sample $Z_0^i$ is then fed into the mask prototype learning branch as a realistic partial observation. 

\subsection{Mask Prototype Learning}
\label{sec:mpl}

Given realistic occluded samples generated by SPT, the core of DPM++ is to perform reliable matching under incomplete observations. 
To this end, we propose \emph{mask prototype learning}, which reformulates occluded ReID as dynamic matching in a prototype space. 
Instead of directly comparing incomplete visual features against holistic identity centers, DPM++ first learns a visual prototype matrix, then predicts an image-specific prototype mask, and finally matches the current input against the corresponding masked prototype subspace. 
Therefore, the matching process becomes visibility-conditioned rather than globally fixed.

Let $f_{cls}\in\mathbb{R}^{1\times c}$ denote the CLS feature extracted by the backbone. 
We maintain a learnable prototype matrix for each identity as:
\begin{equation}
P \in \mathbb{R}^{K\times c},
\label{eq:proto_matrix}
\end{equation}
where $K$ is the number of identities and $c$ is the feature dimension. 
For each input instance, an image-specific prototype mask:
\begin{equation}
M_p \in \mathbb{R}^{1\times c}
\label{eq:proto_mask}
\end{equation}
is predicted and applied to the prototype matrix to form a masked prototype matrix:
\begin{equation}
P_m = P \odot M_p.
\label{eq:masked_proto_full}
\end{equation}
The CLS feature is then jointly supervised by the coarse prototype matrix $P_c$, the learnable prototype matrix $P$, and the masked prototype matrix $P_m$. 
These three spaces respectively encode semantic anchoring, standard visual discrimination, and visibility-consistent subspace matching.

\subsubsection{Early Semantic Anchoring}
As mentioned in the CLIP-ReID~\cite{li2023clip}, using a learning text anchor for person re-identification task will gain a better performance than directly learning from visual space.
Prototype learning under heavy occlusion is prone to bias, because the learned visual class centers are inevitably shaped by incomplete and contaminated observations. 
To alleviate this problem, DPM++ follows the observation of CLIP-ReID~\cite{li2023clip} and introduces \emph{early semantic anchoring}. 
The goal of this design is not to replace the final visual prototype space with textual features, but to provide a semantically informed coarse identity organization before the learnable visual prototype matrix is fully optimized. 
This mechanism serves as an early teacher signal that stabilizes prototype learning under occlusion.

For each identity $y$, we construct an ID-specific textual prompt:
\begin{equation}
t_y = [\texttt{A photo of a } X_1 X_2 \cdots X_M \texttt{ person}],
\label{eq:text_prompt}
\end{equation}
where $X_1,\dots,X_M$ are learnable textual tokens. 
A frozen CLIP text encoder and a frozen CLIP image encoder are used to optimize these prompts through image-text alignment. 
After optimization, all identity-level text embeddings are collected into a \emph{coarse prototype matrix}
\begin{equation}
P_c \in \mathbb{R}^{K\times c}.
\label{eq:coarse_proto_full}
\end{equation}
As shown in Fig.~\ref{fig:overview}, this coarse prototype matrix is frozen in the subsequent stage and is used only for \emph{early anchoring}, i.e., semantic regularization of the learnable visual prototype space.

Given a training sample $i$ with identity label $y_i$, we apply an identity classification loss over the coarse prototype matrix:
\begin{equation}
\mathcal{L}_{id}^{c}
=
-\frac{1}{B}\sum_{i=1}^{B}
\log
\frac{\exp(\langle f_{cls}^{\,i},P_c^{y_i}\rangle)}
{\sum_{k=1}^{K}\exp(\langle f_{cls}^{\,i},P_c^{k}\rangle)},
\label{eq:lid_coarse}
\end{equation}
where $B$ is the batch size. 
This term regularizes the visual representation with semantically informed coarse identity anchors. 
Since $P_c$ is frozen, the coarse branch does not dominate the downstream prototype space, but instead provides a stable identity organization that guides the early evolution of the learnable prototypes.

\subsubsection{Prototype Space and Masked Matching}

The learnable prototype matrix $P$ is optimized to capture task-oriented identity centers in the visual space. 
Given the CLS feature $f_{cls}^{\,i}$, the standard identity classification loss over $P$ is
\begin{equation}
\mathcal{L}_{id}^{p}
=
-\frac{1}{B}\sum_{i=1}^{B}
\log
\frac{\exp(\langle f_{cls}^{\,i},P^{y_i}\rangle)}
{\sum_{k=1}^{K}\exp(\langle f_{cls}^{\,i},P^{k}\rangle)}.
\label{eq:lid_proto}
\end{equation}
This branch preserves the discriminative power of the visual prototype space and plays the same role as standard classification in prototype-based ReID.

However, for occluded inputs, directly matching an incomplete CLS feature against the full prototype matrix remains problematic. 
Therefore, DPM++ introduces masked matching by applying an image-specific mask to the prototype matrix. 
The masked prototype matrix $P_m$ emphasizes dimensions that are estimated to remain reliable under the current visibility condition and suppresses unreliable responses caused by occluders or missing body parts. 

A subtle but important issue arises if the masked branch is optimized with exactly the same objective as the original prototype branch. 
In that case, the mask can easily become a negligible factor during optimization. 
Since the original branch already learns a relatively easy discriminative solution, the masked branch may also achieve a low classification loss even with a nearly trivial mask, e.g., one close to an all-one vector. 
As a result, the network has little incentive to learn a truly selective and visibility-aware mask, and the masked branch may collapse into a redundant replica of the original branch. 
To prevent such degeneration, the masked branch should impose a more aggressive optimization objective than the original branch, so that learning a non-trivial prototype mask becomes necessary for further discrimination. 

For this reason, we employ an angular-margin-based identity loss for the masked branch:
\begin{equation}
\mathcal{L}_{id}^{m}
=
-\frac{1}{B}\sum_{i=1}^{B}
\log
\frac{\exp\left(s\left(\cos(\langle f_{cls}^{\,i},P_m^{y_i}\rangle)+m_a\right)\right)}
{\exp\left(s\left(\cos(\langle f_{cls}^{\,i},P_m^{y_i}\rangle)+m_a\right)\right)+D_{inter}^{\,i}},
\label{eq:lid_masked}
\end{equation}
where
\begin{equation}
D_{inter}^{\,i}
=
\sum_{k=1,k\neq y_i}^{K}
\exp\left(s\cos(\langle f_{cls}^{\,i},P_m^{k}\rangle)\right),
\end{equation}
and $s$ and $m_a$ denote the scaling factor and angular margin, respectively. 
This masked branch explicitly enforces the model to learn identity-discriminative matching within the visibility-consistent subspace selected for each instance.

Following previous works~\cite{chen2021occlude,he2021transreid,wang2022pose}, to further enhance the compactness of the visual feature space, we employ a batch-hard triplet loss:
\begin{equation}
\mathcal{L}_{tri}
=
\frac{1}{B}\sum_{i=1}^{B}
\left[
\|f_i-f_i^{pos}\|_2^2
-
\|f_i-f_i^{neg}\|_2^2
+
m_t
\right]_+,
\label{eq:triplet_full}
\end{equation}
where $f_i^{pos}$ and $f_i^{neg}$ are the hardest positive and negative samples of $f_i$ in the current mini-batch, and $m_t$ is the margin. 
Together, $\mathcal{L}_{id}^{c}$, $\mathcal{L}_{id}^{p}$, $\mathcal{L}_{id}^{m}$, and $\mathcal{L}_{tri}$ shape the feature space from semantic, classification, masked matching, and metric learning perspectives.

\subsubsection{Head Enrich Module}
A strong CLS feature is essential for prototype learning, especially when the input is partially observed. 
However, as shown in Fig.~\ref{fig:ori_heat} (a), the attention heads in the last transformer block exhibit consistently high pairwise similarity, with many off-diagonal entries remaining at a relatively large level. 
This suggests that different heads tend to focus on highly overlapping visual patterns, which limits the diversity of multi-head attention and reduces the richness of the CLS representation. 
Since the CLS token serves as the core feature for prototype matching, such head redundancy is particularly harmful in occluded ReID, where the model should aggregate complementary cues from different visible regions. 
Motivated by this observation, we introduce the \emph{Head Enrich Module} (HEM) to explicitly regularize the correlation structure of multi-head attention and encourage different heads to focus on complementary evidence.
 
Let $q_L^{cls}$ denote the query vector of the class token in the last transformer block and $K_L^{img}$ denote the key matrix of the image patches. 
The attention map between the class token and image patches is
\begin{equation}
A(q_L^{cls},K_L^{img})
=
\operatorname{softmax}
\left(
\frac{q_L^{cls}(K_L^{img})^T}{\sqrt{c/N_h}}
\right),
\label{eq:attn_full}
\end{equation}
where $N_h$ is the number of attention heads. 
To encourage head diversity, we impose an orthogonality constraint on the row-wise normalized attention map:
\begin{equation}
\mathcal{L}_{hem}
=
\left\|
\dot{A}\dot{A}^T - \mathbb{I}_{N_h}
\right\|_F^2,
\label{eq:hem_full}
\end{equation}
where $\dot{A}$ is the row-wise $\ell_2$-normalized attention matrix and $\mathbb{I}_{N_h}$ is the identity matrix. 
This loss drives the actual correlation toward the desired target correlation shown in Fig.~\ref{fig:overview}, thereby enriching the CLS representation used for prototype matching.

\subsubsection{Hierarchical Mask Generator}

A key component of DPM++ is the \emph{Hierarchical Mask Generator} (HMG), which predicts the image-specific prototype mask used in Eq.~(\ref{eq:masked_proto_full}). 
In addition to head redundancy, we observe from Fig.~\ref{fig:ori_heat} (b) that patch-token representations become increasingly similar in deeper layers. 
This phenomenon indicates that the final-layer token features are overly smoothed and thus contain insufficient local discriminability for reliable prototype mask prediction. 
If the mask generator relies only on the deepest features, the predicted mask may become insensitive to fine-grained visibility differences among patches. 
To address this issue, we introduce a \emph{Hierarchical Mask Generator} (HMG), which aggregates features from multiple layers rather than predicting the mask solely from the final-layer representation so as to preserve both high-level semantics and low-/mid-level discriminative details for prototype mask estimation. 

Specifically, Let $f_l \in \mathbb{R}^{h\times w\times c}$ denote the reshaped patch representation of the $l$-th transformer block after removing the class token:
\begin{equation}
f_l
=
\operatorname{Reshape}[x_1^l;x_2^l;\cdots;x_D^l].
\label{eq:reshape}
\end{equation}
We then aggregate hierarchical features from selected layers and feed them into the HMG:
\begin{equation}
M_p
=
\sigma
\Big(
\operatorname{GAP}
\big(
\mathcal{G}\big(G*(f_1,f_2,\dots,f_L)\big)
\big)
\Big),
\label{eq:hmg_full}
\end{equation}
where $\mathcal{G}$ denotes convolutional layers, $\operatorname{GAP}(\cdot)$ denotes global average pooling, $G$ is a binary gate selecting which hierarchical features are used, and $\sigma(\cdot)$ is the fully connected layer attached with a sigmoid activation. 
The resulting mask $M_p$ is then applied to the learnable prototype matrix through Eq.~(\ref{eq:masked_proto_full}). 
Unlike conventional spatial or channel attention mechanisms, which reweight the feature responses of the current sample itself, HMG does not directly act on the input feature map. 
Instead, it predicts a visibility-aware prototype mask from the current image and applies this mask to the prototype space to determine which dimensions should participate in matching. 
In other words, the mask is not used for self-enhancement. 
Instead, the current input estimates its visible identity subspace, and this estimated subspace is imposed on the gallery matrix used for matching. 
Therefore, HMG is an input-aware and matching-oriented module rather than a conventional feature attention module.

\subsection{Training Objectives}
\label{sec:objectives}

The optimization of DPM++ is performed in three stages. 
We first train the saliency-guided patch transfer (SPT) module as an independent preprocessing component. 
After SPT is learned, it is fixed and used to generate realistic occluded samples for the subsequent training of DPM++. 

The purpose of SPT training is to learn a reliable saliency-based decomposition of each image into an identity set and an occlusion set. 
To ensure that this decomposition is aligned with the ReID task, the SPT module is trained with the same identity discrimination and metric learning objectives as the downstream DPM++ model. 
At the same time, an additional budget regularization is introduced to avoid the trivial solution in which all patches are selected. 
Concretely, the budget loss is defined as
\begin{equation}
L_{budget} =
\frac{1}{D}\sum_{d=1}^{D} M_i^d - \rho,
\end{equation}
where $D$ denotes the number of image patches, $M_i^d \in (0,1)$ is the saliency response of the $d$-th patch for sample $i$, and $\rho$ controls the expected retained patch ratio. 
This term encourages the decision matrix to preserve only the most discriminative identity-related patches while suppressing background and occlusion-related regions.

Accordingly, the objective for training SPT is given by:
\begin{equation}
\mathcal{L}_{sps}
=
\mathcal{L}_{id}^{sps}
+
\mathcal{L}_{tri}^{sps}
+
\mathcal{L}_{budget}.
\label{eq:sps_total_final}
\end{equation}

After the SPT module is trained, it is fixed and used to synthesize realistic occluded samples for DPM++. 
The subsequent training of DPM++ focuses on semantically anchored mask prototype learning. 
Specifically, the CLS representation is jointly supervised by the learnable prototype matrix, the masked prototype matrix, and the coarse prototype matrix, while additional triplet learning and head decorrelation regularization are also imposed. 
The overall objective of DPM++ is:
\begin{equation}
\mathcal{L}_{dpm++}
=
\mathcal{L}_{id}^{c}
+
\mathcal{L}_{tri}
+
\alpha \mathcal{L}_{id}^{p}
+
(1-\alpha) \mathcal{L}_{id}^{m}
+
\beta\mathcal{L}_{hem},
\label{eq:dpmpp_obj_final}
\end{equation}
where $\alpha$ controls the trade-off between the standard prototype classification loss and the masked prototype loss, and $\beta$ controls the strength of the HEM regularization term.
It worth noting that when SPT-synthesized samples are used for training DPM++, we further adopt a candidate-identity ignoring strategy. 
For a synthesized image, the target identity and the candidate occlusion source often come from different persons. 
Although the transferred patches are selected as occlusion/background regions, they may still contain residual identity cues from the candidate image. 
Directly using the candidate identity in classification or triplet mining would introduce ambiguous supervision and may force the model to suppress useful identity-related patterns. 
Therefore, we only keep the target identity label for the synthesized sample and ignore the candidate identity during loss computation. 
This makes the transferred patches act purely as realistic occlusion interference, while the supervision remains centered on the target person.

In summary, SPT and DPM++ are optimized sequentially: 
SPT is first trained to learn task-driven patch decomposition, and the learned SPT module is then used to generate realistic occluded samples for training DPM++. 
This design ensures that the augmentation process itself is aligned with the downstream ReID objective and dataset, while keeping the optimization of mask prototype learning focused on semantically anchored visibility-consistent matching.

\section{Experiment}
\subsection{Datasets and Evaluation Protocol}

We evaluate DPM++ on both occluded and holistic person re-identification benchmarks, so as to comprehensively verify its effectiveness under severe occlusion as well as its generalization to standard ReID settings.

\textbf{Occluded-Duke}~\cite{miao2019pose} is a large-scale benchmark specifically designed for occluded person re-identification. Its training set contains 15,618 images of 702 identities. The query set consists of 2,210 occluded images from 519 identities, and the gallery set contains 17,661 full-body images from 1,110 identities. Owing to the severe and diverse occlusion patterns in practical scenes, Occluded-Duke has become one of the most widely used benchmarks for evaluating the robustness of occluded ReID methods.

\textbf{Occluded-REID}~\cite{zhuo2018occluded} contains 2,000 images from 200 identities, where each identity is associated with 5 full-body images and 5 occluded images. Since this dataset is relatively small, following previous works~\cite{gao2020pose,wang2020high}, we use Market-1501 as the training set and perform evaluation on Occluded-REID only. This protocol has been widely adopted in the literature to assess cross-dataset generalization under occlusion.

\textbf{Market-1501}~\cite{zheng2015scalable} is a standard holistic person re-identification benchmark collected from 6 cameras. It contains 12,936 training images of 751 identities, 3,368 query images of 750 identities, and 19,732 gallery images of the same 750 identities. We use this dataset both as a standard holistic benchmark and as the training source for evaluating Occluded-REID, following common practice.

\textbf{DukeMTMC-reID}~\cite{zheng2017unlabeled} is another widely used holistic ReID benchmark. It consists of 36,441 images of 1,812 identities captured by 8 cameras. Among them, 16,522 images of 702 identities are used for training, while the remaining 2,228 query images and 17,661 gallery images correspond to 702 unseen identities. Since Occluded-Duke is constructed based on DukeMTMC-reID, evaluating on both datasets allows us to examine not only occlusion robustness but also the transferability of the proposed method to the standard holistic setting.

\textbf{Evaluation protocol.}
For fair comparison with previous methods, we adopt the standard Cumulative Matching Characteristics (CMC) and mean Average Precision (mAP) as evaluation metrics. Specifically, the CMC curve reports the matching accuracy at different ranks, and we mainly report Rank-1, Rank-5, and Rank-10 accuracies. The mAP metric further evaluates the overall retrieval quality by considering both precision and ranking order. Unless otherwise specified, all reported results follow the evaluation protocols provided by the corresponding datasets and the settings commonly used in prior occluded ReID works~\cite{gao2020pose,wang2020high,tan2024occluded}. 

\textbf{Implementation details.}
We instantiate DPM++ on top of CLIP~\cite{radford2021learning} with a ViT-B/16 visual backbone. 
Following TransReID~\cite{he2021transreid}, the patch stride is set to $11 \times 11$, and the Side Information Embedding (SIE) is adopted with coefficient 1.0. 
The visual embedding dimension is 768, which is further projected to 512 in order to align with the text embedding space. 
All input images are resized to $256 \times 128$. 
Standard data augmentation strategies, including random horizontal flipping, random cropping, and random erasing~\cite{zhong2020random}, are applied during training. 
Each mini-batch contains 64 images sampled from 4 identities. 
Our framework is implemented in PyTorch and trained on four NVIDIA RTX 3090 GPUs.

The training of DPM++ is conducted in three stages, corresponding to semantic prior initialization, saliency-guided patch decomposition, and semantically anchored mask prototype learning, respectively.

\textit{Stage I: Semantic prior initialization.}
In the first stage, we optimize only the ID-specific prompt tokens $[X]_1,\ldots,[X]_M$ to construct the coarse prototype matrix used for early semantic anchoring. 
The CLIP image encoder and text encoder remain frozen in this stage. 
The optimization is performed for 100 epochs using Adam, with a base learning rate of $3.5 \times 10^{-4}$. 
A linear warmup schedule is applied at the beginning of training, followed by cosine learning-rate decay.

\textit{Stage II: SPS pre-training.}
In the second stage, we pre-train the Salient Patch Selection (SPS) module following the setting of SPT~\cite{tan2024occluded}. 
The purpose of this stage is to learn a reliable decomposition of each training image into identity and occlusion sets before the synthesized occluded samples are introduced into the final training pipeline. 
The SPS module is optimized using SGD with an initial learning rate of $8 \times 10^{-3}$. 
Meanwhile, consistent with CLIP-ReID-style optimization, the backbone and the classification heads are updated by Adam with a smaller learning rate of $5 \times 10^{-5}$, together with a 5-epoch linear warmup. 

\textit{Stage III: Training of DPM++.}
After SPS pre-training, the learned SPS module is fixed and used to synthesize realistic occluded samples for the final training of DPM++. 
In this stage, the visual encoder and the mask prototype learning modules are jointly optimized. 
Specifically, the DPM-related modules are trained with Adam using a learning rate of $5 \times 10^{-3}$, while the visual encoder is updated with a smaller learning rate of $1 \times 10^{-5}$. 
The training lasts for 60 epochs. 
We set the weight decay to $10^{-4}$ and adopt a step decay schedule, where the learning rate is multiplied by 0.1 at the 20-th and 40-th epochs.

For the SPT-related hyper-parameters, following~\cite{tan2024occluded}, we set $\beta=0.3$ and $\alpha_1=0.5$. 
For the mask rolling criterion, we set $\alpha_2=0.1$ and adopt the ranking strategy proposed in SPT: within each mini-batch, only the top 10\% samples ranked by the OIoU score after mask rolling are retained as valid candidates for occlusion synthesis.

\begin{table}[t]
  \centering
  \renewcommand{\arraystretch}{1.2}
  \resizebox{\columnwidth}{!}{ 
  \begin{tabular}{l|cc|cc}
  \toprule[1pt]
   \multirow{2}{*}{\textbf{Method}} &\multicolumn{2}{c|}{\textbf{Occluded-Duke}} & \multicolumn{2}{c}{\textbf{Occluded-REID}}\\
   \cline{2-5}
   & R-1 & \emph{m}AP & R-1 & \emph{m}AP\\
  \hline
    PCB \cite{sun2018beyond}                & 42.6 & 33.7 & 41.3 & 38.9 \\
    Part Bilinear \cite{suh2018part}        & 36.9 & -    & -    & -    \\
    FD-GAN \cite{ge2018fd}                  & 40.8 & -    & -    & -    \\
    ISP \cite{zhu2020identity}              & 62.8 & 52.3 & -    & -    \\
    \hline    
    DSR \cite{he2018deep}                   & 40.8 & 30.4 & 72.8 & 62.8 \\
    FPR \cite{he2019foreground}              & -    & -    & 78.3 & 68.0 \\
    PGFA \cite{miao2019pose}                & 51.4 & 37.3 & -    & -    \\
    PVPM~\cite{gao2020pose}       & 47.0 & 37.7 & 66.8 & 59.5 \\
    HOReID~\cite{wang2020high}    & 55.1 & 43.8 & 80.3 & 70.2 \\
    OAMN~\cite{chen2021occlude}   & 62.6 & 46.1 & -    & -    \\
    RFCnet~\cite{hou2021feature}                       & 63.9 & 54.5 & -    & -    \\
    PAT*~\cite{li2021diverse}     & 64.5 & 53.6 & 81.6 & 72.1 \\
    TransReID*~\cite{he2021transreid} & 66.4 & 59.2 & - & - \\
    DRL-Net~\cite{jia2022learning}                        & 65.8 & 53.9 & -    & -    \\
    FED~\cite{wang2022feature}                           & 68.1 & 56.4 & 86.3 & 79.3 \\
    PFD~\cite{wang2022pose}                           & 69.5 & 61.8 & 81.5 & 83.0 \\
    CAAO~\cite{zhao2023content}                           & 68.5 & 59.5 & 87.1 & 83.4 \\
    HCGA~\cite{dou2022human}                           & 70.2 & 57.5 & -    & -    \\
    SAP~\cite{jia2023semi}                           & 70.0 & 62.2 & 83.0 & 76.8 \\
    OAT~\cite{li2024occlusion}                            & 71.8 & 62.2 & 82.6 & 78.2 \\
    ADP~\cite{xia2024attention}                           & 74.5 & 63.8 & 89.2 & 85.1 \\
    \hline
    DPM~\cite{tan2022dynamic}                       & 71.4 & 61.8 & 85.5 & 79.7\\
    CLIP-ReID~\cite{li2023clip}                     & 67.1 & 59.5 & -    & -    \\
    DPM-SPT~\cite{tan2024occluded}                       & 74.7 & 63.0 & 87.8 & 81.1 \\
    \textbf{DPM++ (Ours)}                & \textbf{76.9} & \textbf{67.2} & \textbf{95.4} & \textbf{91.8} \\
  \bottomrule[1pt]
    \end{tabular}}
    \vspace{0em}\caption{\textbf{Comparison with previous state-of-the-art methods on Occluded-Duke and Occluded-REID.} 
    The upper methods are designed for holistic ReID and the lower methods are for occluded ReID. 
    The symbol $*$ represents methods that employ the transformer structure.}
    \label{Tbl:osota}
\end{table}
\begin{table}[t]
  \centering
  \renewcommand{\arraystretch}{1.2}
  \resizebox{0.95\columnwidth}{!}{ 
  \begin{tabular}{l|cc|cc}
  \toprule[1pt]
   \multirow{2}{*}{\textbf{Method}} &\multicolumn{2}{c|}{\textbf{Market-1501}} & \multicolumn{2}{c}{\textbf{DukeMTMC}}\\
   \cline{2-5}
   & R-1 & \emph{m}AP & R-1 & \emph{m}AP\\
   \hline
    PCB \cite{sun2018beyond}                & 92.3 & 71.4 & 81.8 & 66.1 \\
    MGN \cite{wang2018learning}             & 95.7 & 86.9 & 88.7 & 78.4 \\
    ISP \cite{zhu2020identity}              & 95.3 & 88.6 & 89.6 & 80.0 \\
    SAN~\cite{jin2020semantics}                           & 96.1 & 88.0 & 87.9 & 75.5 \\
    TransReID* \cite{he2021transreid}       & 95.2 & 88.9 & 90.7 & 82.0 \\
    HAT*~\cite{zhang2021hat}                     & 95.8 & 89.8 & 90.4 & 81.4 \\
    DCAL~\cite{zhu2022dual}                          & 94.7 & 87.5 & 89.0 & 80.1 \\
    AAformer*~\cite{zhu2023aaformer}                    & 95.4 & 88.0 & 90.1 & 80.9 \\
    PHA~\cite{zhang2023pha}                           & \textbf{96.1} & 90.2 & -    & -    \\
   \hline
    FPR \cite{he2019foreground}             & 95.4 & 86.6 & 88.6 & 78.4 \\
    PGFA \cite{miao2019pose}                & 91.2 & 76.8 & 82.6 & 65.5 \\
    HOReID \cite{wang2020high}              & 94.2 & 84.9 & 86.9 & 75.6 \\
    OAMN \cite{chen2021occlude}             & 93.2 & 79.8 & 86.3 & 72.6 \\
    PAT* \cite{li2021diverse}               & 95.4 & 88.0 & 88.8 & 78.2 \\
    RFCnet~\cite{hou2021feature}                       & 95.2 & 89.2 & 90.7 & 80.7 \\
    PFD~\cite{wang2022pose}                           & 95.5 & 89.7 & 91.2 & 83.2 \\
    FED~\cite{wang2022feature}                           & 95.0 & 86.3 & 89.4 & 78.0 \\
    CAAO~\cite{zhao2023content}                           & 95.3 & 88.0 & 89.8 & 80.9 \\
    HCGA~\cite{dou2022human}                            & 95.2 & 88.4 & -    & -    \\
    SAP~\cite{jia2023semi}                           & 96.0 & 90.5 & -    & -    \\
    ADP~\cite{xia2024attention}                           & 95.6 & 89.5 & 91.2 & 83.1 \\
    OAT~\cite{li2024occlusion}                           & 95.7 & 89.9 & 91.2 & 82.3 \\
   \hline
    DPM~\cite{tan2022dynamic}                         & 95.5 & 89.7 & 91.0 & 82.6 \\
    CLIP-ReID*~\cite{li2023clip}                    & 95.5 & 89.6 & 90.0 & 82.5 \\
    DPM-SPT*~\cite{tan2024occluded}                      & 95.5 & 89.4 & 91.1 & 82.4 \\
    \textbf{DPM++ (Ours)}                & \textbf{96.1} & \textbf{91.2} & \textbf{91.4} & \textbf{83.9} \\
   \bottomrule[1pt]
  \end{tabular}}
  \vspace{0em}\caption{\textbf{Comparison with state-of-the-art methods in terms of CMC (\%) and \emph{m}AP (\%) on Market-1501 and DukeMTMC.} 
  The symbol $*$ represents methods that employ the transformer structure.}
  \label{Tbl:hsota}
\end{table}

\subsection{Comparison with State-of-the-Art Methods}

We compare DPM++ with previous state-of-the-art methods on both occluded and holistic person re-identification benchmarks. 
The comparison includes general holistic ReID methods, occlusion-specific ReID methods, and the closely related baselines of DPM~\cite{tan2022dynamic}, CLIP-ReID~\cite{li2023clip}, and SPT~\cite{tan2024occluded}. 
The results are reported in Tables~\ref{Tbl:osota} and~\ref{Tbl:hsota}.

\textbf{Comparison on occluded ReID benchmarks.}
Table~\ref{Tbl:osota} reports the comparison on Occluded-Duke and Occluded-REID. 
It can be observed that DPM++ achieves the best performance on both datasets by a clear margin, demonstrating its strong robustness under severe occlusion. 
Specifically, on Occluded-Duke, DPM++ achieves \textbf{76.9\%} Rank-1 accuracy and \textbf{67.2\%} mAP, surpassing the previous best competitor ADP by \textbf{2.4\%} in Rank-1 and \textbf{3.4\%} in mAP. 
Compared with our conference predecessor DPM~\cite{tan2022dynamic}, DPM++ improves the Rank-1 accuracy from 71.4\% to 76.9\% and the mAP from 61.8\% to 67.2\%, yielding substantial gains of \textbf{5.5\%} and \textbf{5.4\%}, respectively. 
Moreover, compared with general combination of DPM and SPT~\cite{tan2024occluded} (DPM-SPT), which integrates realistic occlusion synthesis SPT into the original DPM framework, DPM++ still brings further improvements of \textbf{2.2\%} in Rank-1 and \textbf{4.2\%} in mAP, showing that the proposed semantically anchored mask prototype learning is highly complementary to realistic occlusion augmentation.

The performance gain is even more significant on Occluded-REID. 
DPM++ achieves \textbf{95.4\%} Rank-1 accuracy and \textbf{91.8\%} mAP, substantially outperforming all compared methods. 
In particular, compared with the strongest previous competitor ADP~\cite{xia2024attention}, DPM++ improves Rank-1 by \textbf{6.2\%} and mAP by \textbf{6.7\%}. 
Compared with DPM, the gains reach \textbf{9.9\%} in Rank-1 and \textbf{12.1\%} in mAP, while over DPM-SPT the improvements are still as large as \textbf{7.6\%} and \textbf{10.7\%}, respectively. 
Such consistent and pronounced gains on two different occluded benchmarks verify that DPM++ effectively addresses the mismatch between occlusion-corrupted inputs and holistic identity modeling. 
More importantly, these results show that the proposed framework does not merely improve robustness through stronger augmentation, but learns a more reliable visibility-consistent matching mechanism in the prototype space.

Another notable observation is that DPM++ consistently outperforms both holistic ReID methods and occlusion-specific competitors. 
Compared with representative transformer-based ReID baselines such as TransReID~\cite{he2021transreid} and CLIP-ReID~\cite{li2023clip}, the superiority of DPM++ suggests that simply strengthening the backbone or introducing vision-language priors is insufficient for occluded ReID unless the matching process itself is redesigned to explicitly account for visibility inconsistency. 
Similarly, compared with recent occlusion-oriented methods such as SAP, OAT, and ADP, the results indicate that combining realistic occlusion synthesis, early semantic anchoring, and dynamic mask prototype learning provides a more effective solution than relying on alignment or augmentation alone.

\begin{table*}[t]
  \centering
  \renewcommand{\arraystretch}{1.12}
  \resizebox{0.8\textwidth}{!}{
  \begin{tabular}{l|cccc|cccc}
  \toprule[1pt]
  \multirow{2}{*}{\textbf{Configuration}}
  & \multicolumn{4}{c|}{\textbf{CLIP-ViT Backbone}} 
  & \multicolumn{4}{c}{\textbf{ViT Backbone}} \\
  \cline{2-9}
  
  & \textbf{R-1} & \textbf{R-5} & \textbf{R-10} & \textbf{\emph{m}AP}
  & \textbf{R-1} & \textbf{R-5} & \textbf{R-10} & \textbf{\emph{m}AP} \\
  \hline
  Baseline
  & 71.2 & 82.9 & 87.9 & 63.2
  & 60.7 & 77.0 & 82.5 & 53.0 \\
  \hline
  \hline
  + SPT
  & 71.1 & 84.2 & 88.2 & 63.4
  & 65.3 & 80.4 & 85.4 & 55.8 \\

  + SPT+ OIoU
  & 73.3 & 86.7 & 90.1 & 65.2
  & 66.6 & 81.6 & 86.7 & 56.5 \\

  + SPT + OIoU + MR
  & 75.2 & 87.0 & 90.5 & 65.9
  & 68.6 & 82.8 & 87.5 & 57.4 \\
  \hline
  \hline
  +DPM
  & 74.0 & 85.7 & 89.5 & 65.9
  & 70.1 & 82.8 & 86.9 & 59.9 \\

  +DPM+HMG
  & 74.3 & 85.5 & 89.5 & 66.2
  & 71.0 & 82.9 & 87.2 & 61.0 \\

  +DPM+HMG+HEM
  & 74.3 & 85.7 & 89.6 & 66.4
  & 71.4 & 83.7 & 87.4 & 61.8 \\
  \hline
  \hline

  Full Version
  & \textbf{76.9} & \textbf{88.2} & \textbf{90.7} & \textbf{67.2}
  & \textbf{74.7} & \textbf{86.0} & \textbf{89.8} & \textbf{63.0} \\
  \bottomrule[1pt]
  \end{tabular}}
  \caption{\textbf{Cross-backbone ablation study on Occluded-Duke.} 
  The left block reports the results based on the CLIP-ViT backbone, while the right block reports the corresponding results based on the vanilla ViT backbone. 
  SPT denotes Saliency-Guided Patch Transfer, OIoU denotes Occlusion-Aware IoU, MR denotes Mask Rolling, HMG denotes hierarchical mask generator, and HEM denotes the Head Enrich Module.}
  \label{tab:cross_backbone_ablation}
\end{table*}

\textbf{Comparison on holistic ReID benchmarks.}
Table~\ref{Tbl:hsota} reports the comparison on the standard holistic ReID benchmarks Market-1501 and DukeMTMC-reID. 
Although DPM++ is primarily designed for occluded person re-identification, it still achieves highly competitive and even state-of-the-art performance under the standard holistic setting. 
On Market-1501, DPM++ achieves \textbf{96.1\%} Rank-1 accuracy and \textbf{91.2\%} mAP, matching the best reported Rank-1 and achieving the best mAP among all compared methods. 
On DukeMTMC-reID, DPM++ obtains \textbf{91.4\%} Rank-1 accuracy and \textbf{83.9\%} mAP, outperforming all competitors on both metrics.

Compared with orginal DPM~\cite{tan2022dynamic}, DPM++ improves the mAP from 89.7\% to 91.2\% on Market-1501 and from 82.6\% to 83.9\% on DukeMTMC-reID, while also bringing consistent gains in Rank-1 accuracy. 
Compared with CLIP-ReID~\cite{li2023clip}, DPM++ improves the mAP by \textbf{1.6\%} on Market-1501 and \textbf{1.4\%} on DukeMTMC-reID. 
Compared with compared with general combination of DPM and SPT~\cite{tan2024occluded} (DPM-SPT), the gains are also consistent on both datasets. 
These results are important because they indicate that the proposed framework does not overfit to occlusion-specific cases at the expense of standard ReID performance. 
Instead, the learned prototype space remains generally discriminative even when the full person body is visible.

The above comparisons lead to two conclusions. 
First, the largest gains of DPM++ are observed on occluded benchmarks, which confirms that the proposed framework is particularly effective in handling severe visibility conditions. 
Second, DPM++ also preserves strong performance on standard holistic benchmarks, indicating that the proposed semantic anchoring and mask prototype learning do not harm the general discriminative ability of the learned representation. 
Taken together, these results demonstrate that DPM++ provides a unified and effective solution for both occluded and standard person re-identification scenarios.

\subsection{Ablation Study}

To comprehensively evaluate the contribution of each component in DPM++, we conduct a cross-backbone ablation study on Occluded-Duke, as reported in Table~\ref{tab:cross_backbone_ablation}. 
In particular, we perform all ablations on both a CLIP-ViT backbone and a vanilla ViT backbone, so as to examine whether the effectiveness of the proposed components depends on a specific visual encoder. 
The ablation is organized into two groups: the first group evaluates the contribution of the saliency-guided patch transfer branch, including SPT, OIoU, and mask rolling (MR), while the second group evaluates the contribution of the mask prototype learning branch, including DPM, HMG, and HEM. 
The full version corresponds to the complete DPM++ framework that combines all proposed components.

\textbf{Effect of saliency-guided patch transfer.}
We first evaluate the contribution of the realistic occlusion synthesis branch. 
Starting from the CLIP-ViT baseline, directly introducing SPT yields a marginal improvement in mAP from 63.2\% to 63.4\%, while Rank-1 remains nearly unchanged. 
This result indicates that simply decomposing and recombining patches is not yet sufficient to produce consistently beneficial training samples. 
However, once OIoU is introduced, the performance improves more clearly to 73.3\% Rank-1 and 65.2\% mAP, showing that a more suitable identity-occlusion pairing strategy is critical for effective sample construction. 
After further incorporating mask rolling, the performance increases to 75.2\% Rank-1 and 65.9\% mAP, which demonstrates that spatial compatibility and scale consistency between the identity set and the occlusion set are also important for realistic occlusion synthesis.

A similar trend can be observed on the vanilla ViT backbone. 
Compared with the baseline, SPT alone improves Rank-1 from 60.7\% to 65.3\% and mAP from 53.0\% to 55.8\%, already showing clear gains. 
After adding OIoU and MR, the performance further rises to 68.6\% Rank-1 and 57.4\% mAP. 
These consistent gains on both backbones verify that the proposed saliency-guided patch transfer branch indeed provides more effective training samples than naive data augmentation, and that its contribution does not rely on a particular encoder architecture.

\textbf{Effect of mask prototype learning.}
We then analyze the contribution of the mask prototype learning branch. 
On the CLIP-ViT backbone, introducing DPM alone improves the baseline from 71.2\% to 74.0\% in Rank-1 and from 63.2\% to 65.9\% in mAP, showing that prototype-space matching is already more suitable than naive global comparison under occlusion. 
After adding HMG, the mAP further increases from 65.9\% to 66.2\%, which confirms that hierarchical mask estimation provides a more reliable visibility-aware prototype mask than directly relying on the standard feature representation. 
With HEM further introduced, the performance reaches 74.3\% Rank-1 and 66.4\% mAP. 
Although the gain in Rank-1 is relatively small, the improvement in mAP suggests that encouraging more diverse head responses helps produce more discriminative CLS features and benefits overall retrieval quality.

The same pattern is also observed on the vanilla ViT backbone. 
DPM alone improves the baseline by 9.4\% in Rank-1 and 6.9\% in mAP, confirming that dynamic prototype masking is particularly beneficial when the backbone itself is less semantically enriched. 
With HMG, the performance further rises to 71.0\% Rank-1 and 61.0\% mAP, and HEM brings an additional gain to 71.4\% Rank-1 and 61.8\% mAP. 
These results consistently demonstrate that the proposed mask prototype learning framework, together with hierarchical mask estimation and head enrichment, effectively improves visibility-consistent matching under occlusion.

\textbf{Complementarity of SPT and mask prototype learning.}
The full version of DPM++ achieves the best results on both backbones, reaching 76.9\% Rank-1 / 67.2\% mAP on CLIP-ViT and 74.7\% Rank-1 / 63.0\% mAP on vanilla ViT. 
Compared with the strongest SPT-only configuration, the full version still brings gains of 1.7\% Rank-1 and 1.3\% mAP on CLIP-ViT, and 6.1\% Rank-1 and 5.6\% mAP on vanilla ViT. 
Compared with the strongest DPM-only configuration, the full version also yields additional improvements on both backbones. 
These results show that the two branches are highly complementary: SPT improves the quality of the training inputs by synthesizing realistic occluded samples, while mask prototype learning improves the matching mechanism itself by dynamically selecting visibility-consistent prototype subspaces. 
Combining the two leads to a more complete solution than using either branch alone.

Finally, a notable observation from Table~\ref{tab:cross_backbone_ablation} is that all proposed components produce consistent gains on both CLIP-ViT and vanilla ViT. 
This is particularly important because it suggests that 

\begin{table}[t]
  \centering
  \renewcommand{\arraystretch}{1.1}
  \resizebox{\columnwidth}{!}{
  \begin{tabular}{cccc|cccc}
  \toprule[1pt]
   \multicolumn{4}{c|}{\textbf{Setting}} & \multicolumn{4}{c}{\textbf{Occluded-Duke}} \\
   \cline{1-8}
   $P_{n}$ & $F_{n}$ & $P$ & $F$ & R-1 & R-5 & R-10 & \emph{m}AP\\
  \hline
    \checkmark &            &            &            & 73.9 & \textbf{86.2} & 89.4 & 66.3 \\
               & \checkmark &            &            & \textbf{74.3} & 85.7 & 89.6 & \textbf{66.4} \\
    \checkmark & \checkmark &            &            & 73.9 & 85.9 & 89.1 & 66.0 \\
               &            & \checkmark &            & 73.7 & 85.9 & \textbf{89.7} & \textbf{66.4} \\
               &            &            & \checkmark & 73.8 & 85.4 & 89.1 & 66.3 \\
               &            & \checkmark & \checkmark & 73.8 & 85.4 & 89.1 & 66.3 \\
  \bottomrule[1pt]
    \end{tabular}}
    \caption{\textbf{Comparison of different masking positions on Occluded-Duke.} 
$P$ and $F$ denote that the predicted mask is applied to the prototype matrix and the feature representation before $\ell_2$ normalization, respectively, while $P_n$ and $F_n$ denote the corresponding variants after normalization.}
    \label{Tbl:PF_new}
    \vspace{-2em}
\end{table}

\subsection{Discussions}
\textbf{Prototype-side masking versus feature-side masking.}
Table~\ref{Tbl:PF_new} compares different choices of where the predicted mask should take effect. 
Overall, applying the mask to the prototype side is more consistent with the objective of DPM++, since the goal of the mask is not to refine the current feature itself, but to estimate the visible identity subspace and use it to modify the identity prototype for matching. 
In other words, the proposed mask strategy is inherently matching-oriented rather than feature-refinement-oriented.

We also observe that applying the mask on both sides does not bring additional improvement. 
This suggests that two-sided modulation introduces redundant masking and may weaken the discriminative role of the learned mask. 
Moreover, applying the mask before normalization is generally more effective, because it preserves a clearer suppression effect on unreliable dimensions. 
Based on these observations, we apply the mask to the prototype matrix, which gives a more natural and stable design for occluded person re-identification.

\textbf{Effect of the Head Enrich Module.}
The purpose of HEM is to alleviate the redundancy among different attention heads and thereby enrich the representation aggregated by the class token. 
As discussed in Fig.~\ref{fig:ori_heat}(a), the vanilla transformer exhibits a clear tendency that multiple heads produce highly similar attention patterns, which limits the diversity of the information collected by the CLS token and weakens both prototype learning and mask estimation.

To verify the effect of HEM, we visualize the cross-correlation matrix between multi-head attention maps after introducing the proposed module, as shown in Fig.~\ref{fig:ori_heat}. 
Compared with the original backbone, the correlations among different heads are generally reduced after applying HEM, indicating that the attention responses become less redundant and more complementary. 
Although the heads are still not fully decorrelated, the overall reduction in cross-head similarity suggests that the CLS token can aggregate richer and more diverse visual cues from different image patches. 

This observation is consistent with the role of HEM in our framework. 
By explicitly encouraging different heads to focus on different patterns, HEM improves the discriminative quality of the global representation, which in turn benefits both holistic prototype learning and the prediction of a more informative prototype mask. 
Overall, the visualization provides clear evidence that HEM is effective in reducing head redundancy and enhancing the diversity of the learned attention patterns.

\begin{figure}[t]
    \centering
    \includegraphics[width=0.90\columnwidth]{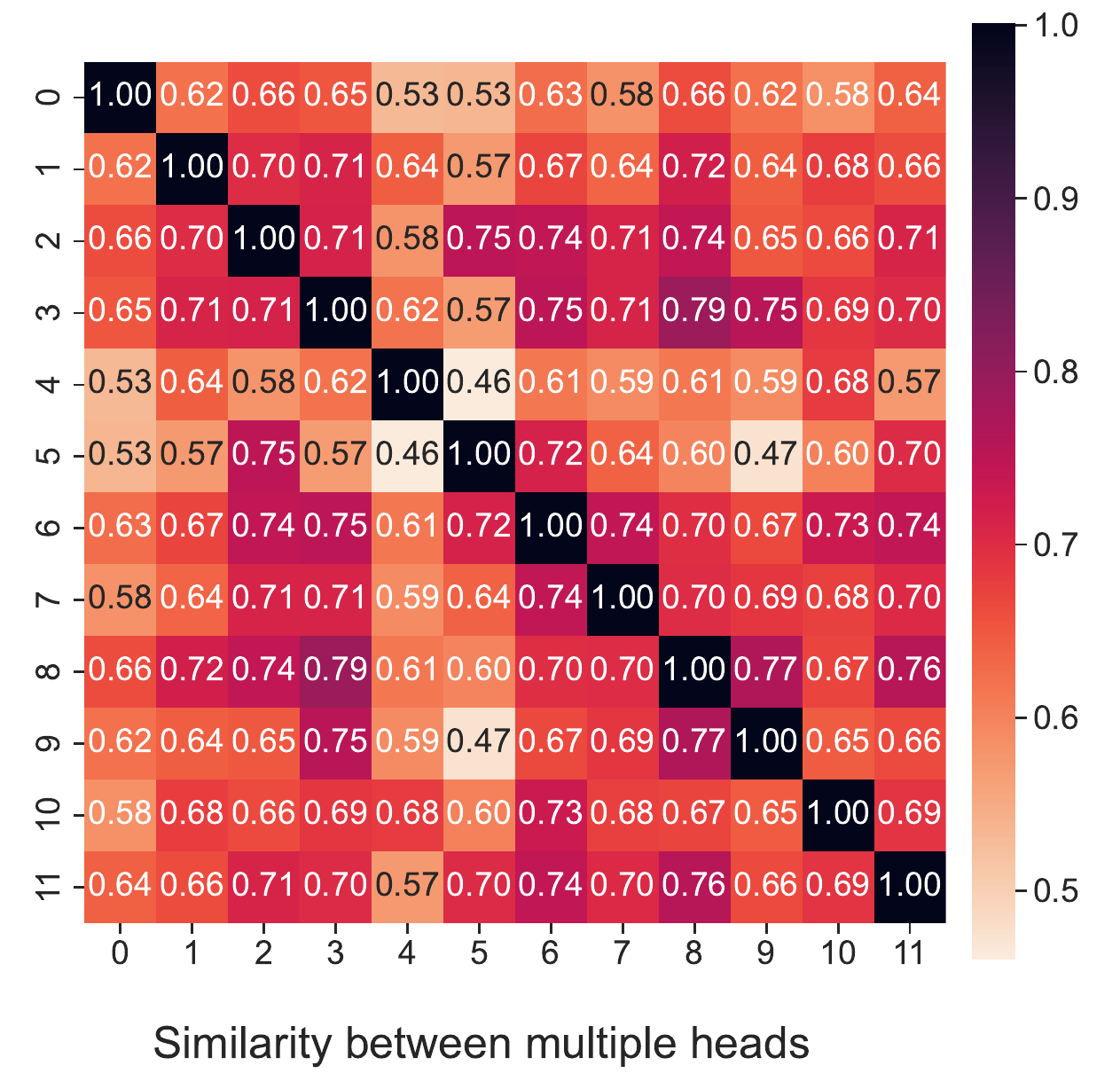}
    \caption{\textbf{Visualization of the cross-correlation matrix between attention maps of different heads after introducing HEM.} Compared with the vanilla transformer shown in Fig.~\ref{fig:ori_heat}(a), the off-diagonal correlations are generally reduced, indicating that HEM alleviates head redundancy and encourages different heads to capture more complementary visual cues. This diversification enriches the CLS representation and benefits both prototype learning and prototype mask estimation.
}
    \label{fig:after_heat}
\end{figure}

\begin{figure}[t]
    \centering
    \includegraphics[width=0.9\columnwidth]{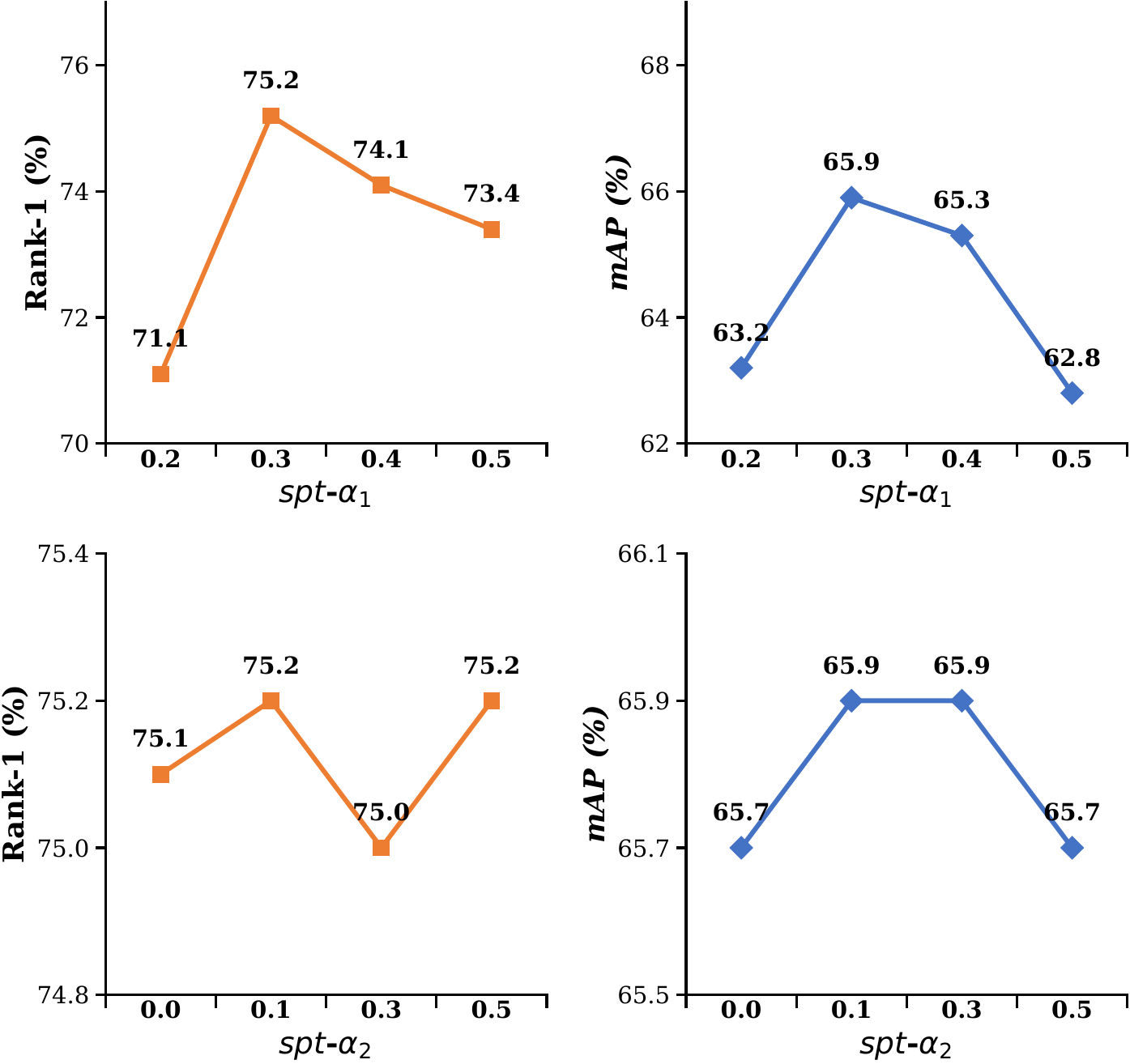}
    \caption{\textbf{Analysis of OIoU threshold $\alpha_1$ and mask-rolling threshold $\alpha_2$ in terms of CMC (\%) and \emph{m}AP (\%) on Occluded-Duke.} A moderate $\alpha_1$ achieves the best trade-off between effective occlusion and sample diversity, while the performance is relatively stable with respect to $\alpha_2$.}
    \label{fig:spt_param}
\end{figure}

\textbf{Analysis of SPT hyper-parameters.}
We further analyze the influence of two key hyper-parameters in SPT, i.e., the OIoU threshold $\alpha_1$ and the mask-rolling threshold $\alpha_2$. 
As shown in Fig.~\ref{fig:spt_param}, $\alpha_1$ has a relatively clear impact on performance. 
When $\alpha_1$ is too small, the selected identity-occlusion pairs may not provide sufficiently effective occlusion, leading to limited gains. 
When $\alpha_1$ becomes too large, the candidate selection becomes overly strict and may reduce the diversity of synthesized occluded samples. 
The best performance is achieved when $\alpha_1=0.3$, obtaining 75.2\% Rank-1 and 65.9\% mAP, which indicates that a moderate OIoU constraint provides a better trade-off between occlusion effectiveness and sample diversity. 
In contrast, the model is less sensitive to $\alpha_2$. 
Across different values of $\alpha_2$, the Rank-1 accuracy and mAP remain relatively stable, suggesting that mask rolling mainly refines spatial compatibility rather than dominating the quality of generated samples. 
Following this observation, we set $\alpha_1=0.3$ and $\alpha_2=0.1$ by default in our experiments.

\begin{table}[t]
  \centering
  \renewcommand{\arraystretch}{1.1}
  \resizebox{\columnwidth}{!}{
  \begin{tabular}{l|cccc}
  \toprule[1pt]
  \textbf{Setting} & \textbf{R-1} & \textbf{R-5} & \textbf{R-10} & \textbf{\emph{m}AP} \\
  \hline
  \multicolumn{5}{l}{\textbf{DPM ($\beta=0.10$)}} \\
  \hline
  DPM ($\alpha=0.0,\beta=0.10$) & 73.1 & 85.1 & 89.5 & 65.4 \\
  DPM ($\alpha=0.5,\beta=0.10$) & 73.1 & 85.3 & 89.0 & 66.1 \\
  DPM ($\alpha=1.0,\beta=0.10$) & \textbf{74.3} & \textbf{85.7} & \textbf{89.6} & \textbf{66.4} \\
  DPM ($\alpha=1.5,\beta=0.10$) & 74.2 & 85.3 & 89.3 & 66.2 \\
  \hline
  \multicolumn{5}{l}{\textbf{DPM ($\alpha=1.0$)}} \\
  \hline
  DPM ($\alpha=1.0,\beta=0.00$) & 73.7 & 85.6 & 89.4 & 65.8 \\
  DPM ($\alpha=1.0,\beta=0.05$) & 73.7 & 85.0 & 89.0 & 66.2 \\
  DPM ($\alpha=1.0,\beta=0.10$) & \textbf{74.3} & \textbf{85.7} & \textbf{89.6} & \textbf{66.4} \\
  DPM ($\alpha=1.0,\beta=0.15$) & 73.1 & \textbf{85.7} & 89.0 & 65.9 \\
  \bottomrule[1pt]
  \end{tabular}}
  \vspace{0em}
  \caption{\textbf{Parameter analysis of DPM on Occluded-Duke.} 
  We evaluate the influence of the balancing coefficients $\alpha$ and $\beta$. 
  In the upper block, $\beta$ is fixed to 0.10 while $\alpha$ varies. 
  In the lower block, $\alpha$ is fixed to 1.0 while $\beta$ varies.}
  \label{Tbl:dpm_param}
\end{table}

\textbf{Impact of the hyper-parameters $\alpha$ and $\beta$.}
We further analyze the influence of two important hyper-parameters in DPM, i.e., $\alpha$ and $\beta$. 
Here, $\alpha$ controls the balance between the standard prototype classification branch and the masked prototype branch, while $\beta$ controls the strength of the head-correlation regularization introduced by HEM. 
Therefore, $\alpha$ determines how much the model emphasizes visibility-consistent masked matching, and $\beta$ determines how strongly different attention heads are encouraged to produce diverse responses.

When studying $\alpha$, we fix $\beta$ to 0.10 and vary $\alpha$ from 0 to 1.5. 
As shown in Table~\ref{Tbl:dpm_param}, the performance gradually improves when $\alpha$ increases from 0 to 1.0. 
Specifically, the mAP improves from 65.4\% to 66.4\%, and the Rank-1 accuracy increases from 73.1\% to 74.3\%. 
This demonstrates that the masked prototype branch provides effective supervision for learning visibility-consistent matching under occlusion. 
However, further increasing $\alpha$ to 1.5 does not bring additional gains, and the performance slightly decreases to 74.2\% Rank-1 and 66.2\% mAP. 
This indicates that an overly large weight on the masked branch may disturb the balance between learning holistic identity prototypes and learning image-specific prototype masks. 
Therefore, we set $\alpha=1.0$ in the following experiments.

We then study the effect of $\beta$ by fixing $\alpha$ to 1.0. 
As shown in Table~\ref{Tbl:dpm_param}, introducing HEM regularization consistently improves the retrieval performance when compared with $\beta=0$. 
The best overall performance is achieved when $\beta=0.10$, yielding 74.3\% Rank-1 accuracy and 66.4\% mAP. 
When $\beta$ is too small, the regularization is insufficient to effectively reduce redundancy among attention heads. 
When $\beta$ becomes too large, the diversity constraint may over-regularize the attention responses and slightly weaken identity discrimination. 
Considering both Rank-1 accuracy and mAP, we set $\beta=0.10$ as the default setting for DPM++.

\begin{table}[t]
  \centering
  \renewcommand{\arraystretch}{1.1}
  \resizebox{0.9\columnwidth}{!}{
  \begin{tabular}{l|cccc}
  \toprule[1pt]
  \multirow{2}{*}{\textbf{Setting}} 
  & \multicolumn{4}{c}{\textbf{Occluded-Duke}} \\
  \cline{2-5}
  & R-1 & R-5 & R-10 & \emph{m}AP \\
  \hline
  w/o Mask              & 75.7 & 87.3 & 90.4 & 66.1 \\
  Visual Prototype      & \textbf{76.9} & \textbf{88.2} & 90.7 & \textbf{67.2} \\
  Text Prototype        & 76.2 & 88.0 & \textbf{91.1} & \textbf{67.2} \\
  \bottomrule[1pt]
  \end{tabular}}
  \vspace{0em}
  \caption{
  \textbf{Comparison of different prototype spaces for applying the predicted mask on Occluded-Duke.}
  'w/o Mask' denotes that no prototype mask is applied. 
  'Visual Prototype' denotes applying the mask to the learnable visual prototype matrix, while 'Text Prototype' denotes applying the mask to the coarse text prototype matrix.
  }
  \label{Tbl:mask_proto_target}
\end{table}

\textbf{Which prototype should be masked?}
We further investigate where the predicted prototype mask should be applied. 
As shown in Table~\ref{Tbl:mask_proto_target}, removing the mask leads to 75.7\% Rank-1 and 66.1\% mAP, which is clearly lower than the masked variants. 
This verifies that explicitly selecting a visibility-consistent prototype subspace is beneficial for occluded ReID.

When the mask is applied to the learnable visual prototype matrix, the model achieves the best Rank-1 accuracy of 76.9\% and the best mAP of 67.2\%. 
Applying the mask to the coarse text prototype matrix also obtains a competitive mAP of 67.2\%, but its Rank-1 accuracy is lower than that of visual prototype masking. 
This suggests that text prototypes provide useful semantic priors, but they are relatively coarse and less task-adaptive than the visual prototypes learned under ReID supervision. 
Therefore, directly performing masked matching on the text prototype space may not fully exploit the discriminative structure required by fine-grained person retrieval.

These results support our design choice: the coarse text prototype matrix is better used as an early semantic anchor, while the final prototype mask should be applied to the learnable visual prototype matrix. 
In this way, DPM++ can benefit from the holistic semantic organization introduced by text priors, while preserving the task-oriented discrimination of the visual prototype space for final matching.

\begin{table}[t]
  \centering
  \renewcommand{\arraystretch}{1.1}
  \resizebox{\columnwidth}{!}{
  \begin{tabular}{l|cccc}
  \toprule[1pt]
   \multirow{2}{*}{\textbf{Method}} & \multicolumn{4}{c}{\textbf{Occluded-Duke}} \\
   \cline{2-5}
   & R-1 & R-5 & R-10 & \emph{m}AP\\
  \hline
    Random Erasing~\cite{zhong2020random}       & 71.2 & 82.9 & 87.9 & 63.1 \\
    SPT                 & 75.0 & 86.8 & 90.0 & 63.6 \\
    Random Erasing~\cite{zhong2020random} + SPT                & \textbf{75.2} & \textbf{87.0} & \textbf{90.5} & \textbf{65.9}\\
  \bottomrule[1pt]
    \end{tabular}}
    \vspace{0em}\caption{\textbf{Ablation study of Random Erasing and SPT module.} The results demonstrate the complementarity between data augmentation and the Saliency-Guided Patch Transfer (SPT).}
    \label{Tbl:spt_ablation}
\end{table}

\textbf{Compatibility with Random Erasing.}
We further study whether SPT is compatible with the commonly used Random Erasing augmentation~\cite{zhong2020random}. 
Different from conventional augmentation methods that perturb an existing image in place, SPT can be regarded as a sample synthesis strategy: it transfers identity and occlusion patches between training samples and generates a new occluded image with realistic scene content. 
Therefore, SPT does not conflict with standard image-level augmentation strategies such as Random Erasing, since they operate at different levels.

As shown in Table~\ref{Tbl:spt_ablation}, using SPT alone clearly improves Rank-1 accuracy from 71.2\% to 75.0\%, demonstrating the benefit of synthesizing realistic occlusion patterns. 
When combined with Random Erasing, the performance is further improved to 75.2\% Rank-1 and 65.9\% mAP. 
This indicates that the two strategies are complementary: Random Erasing enhances robustness to generic local corruption, while SPT provides newly synthesized occluded samples with structured and scene-consistent obstacles. 
Their combination exposes the model to both generic missing-region perturbations and realistic occlusion cases, leading to the best overall performance.

\section{Conclusion}

In this paper, we presented \textbf{DPM++}, a Dynamic Masked Metric Learning framework for occluded person re-identification. 
Instead of relying on extra body-part cues or directly comparing occlusion-corrupted features in a global embedding space, DPM++ reformulates occluded ReID as a visibility-consistent prototype matching problem. 
Specifically, it learns an input-adaptive prototype mask to select reliable identity subspaces from the learnable classifier-prototype space, while a CLIP-based early semantic anchoring strategy provides coarse semantic priors to make the prototypes more holistic and robust. 
To further strengthen the learning of masked matching, we integrate saliency-guided patch transfer to synthesize realistic occluded samples from real training images.
Extensive experiments on both occluded and holistic ReID benchmarks demonstrate that DPM++ consistently outperforms previous state-of-the-art methods and maintains strong generalization under standard ReID settings. 
Ablation studies further verify the effectiveness of our designs. 
We hope DPM++ can offer a useful perspective for occluded ReID by shifting the focus from explicit visible-part localization to input-adaptive matching.

\ifCLASSOPTIONcompsoc
  \section*{Acknowledgments}
  \label{acknowlegement}
This work was supported by the National Science Fund for Distinguished Young Scholars (No.62525605), the National Natural Science Foundation of China (No. U25B2066, No. U22B2051, No.62576300).
\fi

\ifCLASSOPTIONcaptionsoff
  \newpage
\fi

\bibliographystyle{IEEEtran}
\bibliography{IEEEabrv,IEEEfull}

\end{document}